\documentclass[11pt]{article}
\usepackage[utf8]{inputenc}
\usepackage[margin=1in]{geometry}
\usepackage{amsmath}
\usepackage{amsfonts}
\usepackage{amssymb}
\usepackage{graphicx}
\usepackage{hyperref}
\usepackage{enumitem}
\usepackage{titlesec}
\usepackage{xcolor}
 \usepackage[backend=biber, style=numeric, maxbibnames=11, maxcitenames=11, sortcites]{biblatex}

\usepackage[most]{tcolorbox}

\tcbset {
  base/.style={
    arc=0mm, 
    bottomtitle=0.5mm,
    boxrule=0mm,
    colbacktitle=black!10!white, 
    coltitle=black, 
    fonttitle=\bfseries, 
    left=2.5mm,
    leftrule=1mm,
    right=3.5mm,
    title={#1},
    toptitle=0.75mm, 
  }
}

\definecolor{brandblue}{rgb}{0.34, 0.7, 1}
\newtcolorbox{mainbox}[1]{
  colframe=blue!60!black, 
  base={\color{blue!60!black} #1}
}

\newtcolorbox{subbox}[1]{
  colframe=black!30!white,
  base={#1}
}

\titleformat{\section}{\Large\bfseries\color{blue!60!black}}{\thesection}{1em}{}
\titleformat{\subsection}{\large\bfseries\color{blue!40!black}}{\thesubsection}{1em}{}
\titleformat{\subsubsection}{\normalsize\bfseries\color{blue!30!black}}{\thesubsubsection}{1em}{}

\setlist[itemize,1]{label=\textbullet, leftmargin=*}
\setlist[itemize,2]{label=\textendash, leftmargin=*}
\setlist[itemize,3]{label=\textasteriskcentered, leftmargin=*}

\title{\textbf{GenAI-based Multi-Agent Reinforcement Learning towards Distributed Agent Intelligence: \\
A Generative-RL Agent Perspective}}
\author{Hang Wang and Junshan Zhang \\
Department of Electrical and Computer Engineering\\ University of California, Davis}
\date{\today}

\begin{document}

\maketitle

\tableofcontents
\newpage

\section{Vision}


The intersection of two transformative technological trends --- generative artificial intelligence (GenAI) and multi-agent reinforcement learning (MARL) \cite{zhang2021multi} --- is poised to  revolutionize how    distributed intelligence can be achieved in  multi-agent systems. In general, multi-agent learning systems find widespread applications across diverse engineering domains \cite{canese2021multi}, such as autonomous driving in intelligent transportation systems \cite{wang2025ego} and swarm robotics in exploration missions \cite{huttenrauch2019deep}, where complex interactions between multiple intelligent agents must be carefully orchestrated to accomplish their missions that transcend individual capabilities.

These systems are typically modeled as multi-agent partially observed Markov Decision Processes (POMDPs) \cite{sutton1998reinforcement,bertsekas2021multiagent}, which capture the essential characteristics of real-world distributed intelligence scenarios. However, these mathematical formulations unfortunately often suffer from several long-standing theoretical and practical challenges that have hindered progress for decades. More specifically, the exponential growth of state-action spaces as the number of agents increases, the inherently non-stationary nature of multi-agent environments where every agent's learning affects others'  environments \cite{foerster2016learning}, and the fundamental partial observability that characterizes real-world agent interactions have collectively created formidable barriers to achieving truly intelligent distributed systems.

In this paper, we advocate a vision to reconceptualize distributed  agent intelligence in multi-agent systems. Rather than treating agents as isolated policy optimizers that respond `reactively' to environmental stimuli, as in the conventional paradigm that has dominated MARL research,  we envision agents as sophisticated generative models capable of synthesizing complex multi-agent system dynamics and making \textit{`proactive' decisions} based on predictive understanding of future states and interactions. This paradigm shift {\bf from reactive to proactive decision making} represents a fundamental departure from traditional approaches and opens unprecedented possibilities for achieving distributed agent intelligence.

\subsection{The Promise of Generative Multi-Agent Systems: Generative AI as the Catalyst}

   


Generative AI emerges as a crucial catalyst for realizing this transformative vision. The recent breakthroughs in generative modeling, particularly in natural language processing and computer vision~\cite{kingma2013auto,doersch2016tutorial}, have demonstrated the remarkable capacity of these models to understand, predict, and generate complex patterns in high-dimensional data. When applied to multi-agent systems, these capabilities translate into profound new possibilities for agent intelligence and coordination.

The emergence of reasoning and proactive decision-making capabilities in generative-RL agents represents a qualitative leap beyond traditional reactive systems. These agents can anticipate future scenarios, understand the likely consequences of their actions in multi-agent contexts, and make decisions that account for the strategic nature of multi-agent interactions. Unlike conventional agents that simply map current observations to immediate actions, generative-RL agents maintain rich internal models of the environment and other agents, enabling them to engage in sophisticated reasoning  and strategic planning.

We envisage that generative-RL agents  can anticipate environmental changes by modeling the complex dynamics of multi-agent systems, coordinate seamlessly with other agents through enhanced communication and understanding \cite{sukhbaatar2016learning}, and adapt dynamically to evolving multi-agent scenarios. Generative models enable richer agent-to-agent communication by learning meaningful representations that can be shared efficiently, facilitating information exchange about intentions, capabilities, and plans. This enhanced communication capability helps address one of the fundamental challenges in multi-agent coordination: the difficulty of conveying complex internal states and intentions through limited communication channels \cite{wang2023distributed}.

Furthermore, generative models facilitate the sharing of multi-agent dynamics information in ways that were previously impossible. Traditional approaches require agents to explicitly communicate (compressed versions of ) specific state information or action plans, but generative models can learn to share latent representations that capture the essence of complex multi-agent interactions in compact, meaningful forms \cite{jiang2018learning}. This capability enables a new level of collective intelligence where generative-RL agents can move beyond individual optimization to achieve emergent behaviors that truly represent the collective wisdom of the multi-agent system.

\section{Limitations of Conventional MARL Approaches: The Three Curses and Beyond}

It is well known that multi-agent reinforcement learning faces several fundamental challenges that have ``persisted'' despite decades of intensive research effort across the  scientific community. These challenges are so pervasive and deeply-rooted that they have come to be known as ``curses'' in the field, reflecting their seemingly intractable nature; and
the inherent complexity of MARL presents substantial barriers to its practical deployment.

\subsection{The Curse of Dimensionality in Multi-Agent State-Action Spaces}

The dimensionality challenge in MARL manifests as the exponential growth of joint action-state spaces with the number of agents, creating computational and learning challenges that scale catastrophically with system size. Consider a multi-agent system with $n$ agents, where each agent $i$ has an action space of size $|\mathcal{A}_i|$ and an observation space of size $|\mathcal{O}_i|$. The joint action space scales as $\prod_{i=1}^n |\mathcal{A}_i|$, which grows exponentially with the number of agents. Even for relatively modest numbers of agents with moderate individual action spaces, this joint space quickly becomes computationally intractable.

This notorious scaling has profound implications that extend far beyond simple computational complexity \cite{barto2003recent}. The vast joint space requires exponentially more samples to ensure adequate exploration, making sample-efficient learning extremely challenging \cite{moerland2023model}. Traditional tabular methods become completely infeasible for any realistic multi-agent system, while function approximation methods struggle to capture the complex interactions between agents without overfitting or failing to generalize. The curse of dimensionality also affects the representation of value functions and policies, as these must somehow capture the intricate dependencies between agents' actions and the resulting system dynamics \cite{qu2020scalable,zhang2021multi}.

The problem is further compounded by the fact that the effective dimensionality is not just the product of individual agent spaces, but must also account for the complex interaction patterns between agents  \cite{de2020independent,matignon2012independent}. In many real-world scenarios, agents' actions are not independent, and the value of a particular joint action depends on subtle coordination patterns that are difficult to learn and represent efficiently. Traditional approaches that attempt to decompose the problem or use factorized representations often fail to capture these critical interaction patterns, leading to suboptimal coordination and performance \cite{xu2022mingling}.

Generative models offer a promising avenue for addressing the curse of dimensionality through their ability to learn compact, low-dimensional representations of complex high-dimensional data. Recent advances in generative modeling, such as variational autoencoders (VAEs) \cite{kingma2013auto} and world models \cite{ha2018world}, have demonstrated remarkable success in learning meaningful latent representations that capture the essential structure of complex systems while dramatically reducing dimensionality. When applied to multi-agent systems, these techniques can potentially learn representations that capture the essential patterns of multi-agent interactions in low-dimensional latent spaces, making learning and planning computationally tractable even for a multi-agent system with large numbers of agents.

\subsection{The Curse of Non-Stationarity}


The non-stationarity problem in multi-agent systems represents perhaps the most fundamental challenge to applying traditional reinforcement learning techniques~\cite{oliehoek2008optimal,kumar2020conservative}. In single-agent reinforcement learning, the environment is assumed to be stationary from the agent's perspective, meaning that the transition dynamics model and reward functions remain constant over time. This assumption is fundamental to most theoretical guarantees in reinforcement learning and underlies the convergence proofs for many standard algorithms.

However, in multi-agent environments, this stationarity assumption no longer holds unfortunately. Each agent's policy evolution affects the effective environment of all other agents \cite{moerland2023model,silver2017predictron}, creating a complex web of inter-dependencies where every agent faces a continuously changing environment~\cite{lowe2017multi}. As agents update their policies based on their learning experiences, they alter the transition dynamics and effective reward functions that other agents experience on the fly. This creates a moving target problem where the optimal policy for any agent is constantly shifting as other agents adapt their behaviors.

The non-stationarity problem manifests in several critical ways that undermine traditional learning approaches. Policy drift occurs as agents update their policies, causing the effective transition dynamics and reward functions to change for other agents in ways that can invalidate previously learned knowledge \cite{bloembergen2015evolutionary,hernandez2017survey}. Traditional convergence guarantees no longer hold when all agents are simultaneously learning, as the fundamental assumptions underlying these guarantees are violated. The temporal inconsistency problem means that actions that were optimal at time $t$ may become suboptimal at $t+1$ due to changes in other agents' policies, making it difficult for agents to build stable, reliable behavioral strategies.

This non-stationarity creates a complex feedback loop where agents must not only learn about their environment but also continuously adapt to the changing behaviors of other agents \cite{papoudakis2019dealing,hernandez2017survey}. Traditional approaches often struggle with this dynamic, leading to unstable learning, poor convergence properties, and suboptimal final performance \cite{zhang2021multi,gronauer2022multi}. Some approaches \cite{he2016opponent,busoniu2008comprehensive} attempt to model other agents explicitly, but these suffer from the computational complexity of maintaining and updating models of multiple learning agents simultaneously.

Generative models offer a promising solution to the non-stationarity problem by enabling agents to anticipate other agents' policy evolution rather than simply reacting to it. By learning to predict how other agents' behaviors will change over time, generative-RL agents can adapt proactively to the changing multi-agent dynamics. This predictive capability allows agents to make decisions that account for expected future changes in other agents' behaviors, potentially leading to more stable learning and better long-term performance.

\subsection{The Curse of Partial Observability}


Real-world multi-agent systems often operate under conditions of incomplete information, where agents have fundamentally limited visibility into others' states, intentions, and capabilities \cite{zhu2022survey,zhang2021multi}. This partial observability severely constrains coordination effectiveness and decision-making quality, as agents must make critical decisions based on incomplete and often ambiguous information about the system state and other agents' plans.

The partial observability problem in multi-agent systems is particularly challenging because it affects multiple layers of the decision-making process \cite{jaakkola1994reinforcement,omidshafiei2017deep}. At the most basic level, agents cannot directly observe other agents' internal states, including their current observations, beliefs, and intentions. This hidden state uncertainty makes it difficult for agents to predict other agents' future actions or to coordinate effectively. Physical and bandwidth limitations further restrict information sharing, as real-world systems cannot support unlimited communication between agents. Even when communication is possible, agents must decide what information to share, when to share it, and how to interpret the information they receive from others.

The\textit{ communication constraints} in real-world systems add another layer of complexity to the partial observability problem \cite{zhu2024survey}. Agents must operate under limited bandwidth, intermittent connectivity, and potential communication failures. They must also deal with the fact that communication itself can be costly, both in terms of energy consumption and potential information leakage to adversaries. This creates a complex optimization problem where agents must balance the benefits of information sharing against the costs and risks of communication.

\textit{Coordination failures} emerge naturally from insufficient information about teammates' plans and capabilities \cite{zhang2013coordinating}. When agents cannot accurately predict what their teammates will do, they cannot effectively coordinate their actions, leading to conflicts, duplicated efforts, and missed opportunities for cooperation. The problem is exacerbated by the fact that coordination often requires precise timing and sequencing of actions, which is difficult to achieve without detailed information about other agents' states and plans.

Generative models offer powerful tools for addressing partial observability by learning to infer hidden states and generate plausible scenarios for other agents' behaviors. These models can learn from patterns in observable data to make informed predictions about unobservable aspects of the system state. By generating multiple plausible scenarios for how other agents might behave, generative models enable agents to make decisions that are robust to uncertainty about other agents' intentions and capabilities.

\subsection{Additional Challenges}


Beyond the three curses noted above, MARL faces several additional systemic challenges that complicate the development of effective multi-agent learning systems. The \textit{credit assignment problem} represents a fundamental challenge in determining individual agent contributions to collective outcomes \cite{sutton1984temporal}. In sparse reward environments \cite{vecerik2017leveraging} where feedback is delayed and ambiguous, it becomes extremely difficult to determine which agents' actions contributed to positive or negative outcomes. This problem is further complicated by the fact that the value of an agent's actions often depends critically on the actions taken by other agents, making it difficult to assess individual contributions independently.

\textit{Communication overhead} presents another significant challenge, as agents must balance the benefits of information sharing against communication costs and bandwidth constraints~\cite{kim2020communication}. Effective coordination often requires extensive information sharing, but communication resources are typically limited in real-world systems. Agents must carefully optimize what information to share, when to share it, and how to compress and encode information for efficient transmission. The communication problem is further complicated by the need to maintain privacy and security, as sharing too much information can make agents vulnerable to adversarial attacks.

\textit{Agent heterogeneity} adds yet another layer of complexity to multi-agent learning systems \cite{kapetanakis2003reinforcement,gronauer2022multi}. In real-world applications, agents often have different capabilities, objectives, and constraints. Some agents may be more capable or have access to better sensors, while others may have different mission objectives or operational constraints. Managing this heterogeneity requires coordination algorithms that can account for these differences and allocate tasks and responsibilities appropriately. The challenge is compounded by the fact that agent capabilities and objectives may change over time, requiring adaptive coordination strategies.

\textit{Generalization Failures in Dynamic Environments.} Recent evaluations have exposed critical generalization limitations in state-of-the-art MARL methods when faced with environmental variations. A particularly striking example is the performance degradation of leading algorithms like QMIX \cite{rashid2020monotonic} and MAPPO \cite{yu2022surprising} when evaluated on SMACv2 \cite{ellis2212smacv2}, an improved benchmark designed to address the limitations of the original SMAC environment \cite{samvelyan2019starcraft}. While these methods achieved near-perfect performance on the original SMAC benchmark, SMACv2's introduction of procedurally generated scenarios, varied terrain configurations, diverse unit compositions, and enhanced partial observability challenges revealed their fundamental brittleness \cite{ellis2212smacv2}. The benchmark specifically requires algorithms to generalize to previously unseen settings during evaluation, exposing how conventional approaches that rely on memorized coordination patterns fail when environmental distributions shift. This generalization gap highlights a critical limitation: traditional reactive MARL methods learn environment-specific coordination heuristics rather than developing principled understanding of multi-agent interaction dynamics. These failures underscore the inadequacy of current approaches for real-world deployment, where agents must adapt to novel scenarios, changing environmental conditions, and unforeseen agent compositions without extensive retraining \cite{hu2021rethinking}.

\subsection{Limitations of Conventional MARL Approaches}

\subsubsection{Reactive Decision Making}


Traditional multi-agent reinforcement learning approaches suffer from fundamental limitations that stem from their reactive nature and their failure to account adequately for the complex dynamics of multi-agent interactions \cite{canese2021multi}. Conventional approaches typically rely on direct observation-to-action mappings that constitute basic stimulus-response mechanisms \cite{nguyen2020deep}. These \textit{reactive learning paradigms} often suffer from myopic optimization, where agents focus on rewards without sufficient consideration of long-term strategic implications or the effects of their actions on the broader multi-agent system \cite{albrecht2024multi}.

Furthermore, the \textit{local rationality assumption} underlying many traditional approaches leads to individual agent optimization without adequate consideration of system-wide effects and emergent behaviors \cite{yang2020overview,zhang2021multi}. This narrow focus on individual performance can lead to suboptimal collective outcomes, as agents may pursue locally optimal strategies that are globally suboptimal. The lack of anticipatory capabilities in reactive learning approaches indicate that agents cannot predict and prepare for future scenarios, limiting their ability to engage in strategic planning and proactive coordination \cite{shoham2003multi}.

\subsubsection{Coordination Challenges}


\textit{Coordination challenges} in conventional approaches stem from heavy reliance on implicit coordination mechanisms that depend on trial-and-error learning to discover effective coordination patterns \cite{canese2021multi,du2021survey}. This approach is inefficient and often fails to discover optimal coordination strategies, particularly in complex environments with many possible coordination patterns. \textit{Communication bottlenecks} arise from limited bandwidth and simplistic communication protocols that fail to convey the rich information needed for effective coordination~\cite{pretorius2020learning}. \textit{Scalability} issues plague traditional approaches, as their performance typically degrades significantly with increasing numbers of agents due to the exponential growth in complexity and the inability to manage complex multi-agent interactions effectively.

\section{GenAI Models as Foundations for Generative Multi-Agent RL}

The fundamental innovation in our proposed  GenAI-based approach lies in leveraging generative models as foundational building blocks for next-generation multi-agent reinforcement learning systems. This represents a new learning paradigm   where both environment dynamics and action policies are viewed as generative models, enabling sophisticated prediction, coordination, and decision-making capabilities that are previously impossible with traditional approaches.

The core insight driving this transformation is that generative models possess inherent capabilities for understanding, predicting, and generating complex patterns in high-dimensional data \cite{cao2024survey,harshvardhan2020comprehensive,salakhutdinov2015learning}. When applied to multi-agent systems, these capabilities translate into agents that can synthesize realistic multi-agent scenarios, predict the behavior of other agents, and generate sophisticated coordination strategies \cite{awais2025foundation,qin2024tool}. This generative approach enables agents to move beyond reactive decision-making and engage in proactive planning and strategic reasoning that accounts for the complex dynamics of multi-agent interactions.

\subsection{Environment Dynamics via Generative Modeling}

\subsubsection{World Models}


The modeling of environmental dynamics represents a fundamental departure from traditional approaches that rely either on analytical models or  function approximation. Instead, we propose using sophisticated generative models that can learn to represent and predict the complex dynamics of multi-agent environments through data-driven approaches that capture both the underlying physical laws and the emergent patterns that arise from multi-agent interactions \cite{moerland2023model,polydoros2017survey,berk2019exploiting} \cite{ha2018world,hafner2019learning,hafner2020mastering}.World model based  RL is emerging as  a promising paradigm to improve sample efficiency by enabling agents to exploit a learned  model for the physical environment   (see, e.g., \cite{moerland2023model,polydoros2017survey,berk2019exploiting} \cite{ha2018world,hafner2019learning,hafner2020mastering}).  Different from conventional approaches, world-model based RL takes  an {\em end-to-end learning} approach, where the building blocks (such as dynamics model  and action policy) are  trained and optimized to achieve a single overarching goal.   Recent studies show that world model \cite{ha2018world} based agents have exhibited state-of-the-art performance on a wide range visual control tasks, such as Atari benchmark \cite{bellemare2013arcade}, Deepmind Lab tasks \cite{beattie2016deepmind} and Minecraft game \cite{duncan2011minecraft}. 
 In particular, \cite{ha2018world} uses Variational Autoencoders (VAE) \cite{doersch2016tutorial,kingma2013auto} in the Vision Module to encode each sensory input.  Dreamer \cite{hafner2020mastering,hafner2023mastering} applies convolutional neural networks (CNN) \cite{lecun1989backpropagation} to encode   the hidden state, 
 and use the recurrent models to facilitate agents to be predictive of the future and plan their actions accordingly \cite{levine2013guided,wang2019benchmarking,zhu2020bridging}.   PlaNet \cite{hafner2019learning} proposed Recurrent State Space Model (RSSM) \cite{hafner2023mastering,hafner2020mastering}  to predict the future state and reward and use model predictive control (MPC) \cite{garcia1989model} to adapt the agent's plan during online interaction. 
 {\em Going beyond the single agent world models,  there are few recent studies for the multi-agent case.} \cite{krupnik2020multi,pan2022iso} mainly focus on centralized setting \cite{zhang2021multi, CWM-du2025}, in which a world model is learnt to  control all agents through disentangled representation learning. \cite{wang2023leveraging} considers centralized training and decentralized  execution (CTDE) \cite{oliehoek2008optimal,lowe2017multi} while assuming all the agents share the same reward. In contract, this project studies a general setting for distributed  agent intelligence.

World models have great potential to serve as the cornerstone of this approach, representing generative models that capture multi-agent environment dynamics in ways that enable both reasoning  and prediction. These models learn environment representations that go beyond simple state estimation to capture the rich temporal and spatial patterns that characterize multi-agent systems. Unlike traditional models that may focus on individual agent dynamics, world models for multi-agent systems can capture the complex interaction patterns between agents and how these interactions influence environment evolution over time.

The e\textit{nvironment representation }learned by world models encompasses not just the physical state of the environment but also the patterns of agent interactions and their effects on environment dynamics. These models learn to represent how agents collectively influence environment evolution through their joint actions, capturing both direct effects and indirect consequences that may emerge from complex interaction patterns. Furthermore,  the scalable simulation capability of these models enables efficient generation of realistic multi-agent scenarios that can be used for training, evaluation, and planning purposes.

\textit{Agent interaction modeling} represents a particularly sophisticated aspect of world model development, as these models must learn to capture how different combinations of agent actions lead to different environmental outcomes. This requires understanding not just individual agent capabilities but also the synergistic and conflicting effects that arise when multiple agents act simultaneously in the same environment. The models must learn to represent how agent interactions can lead to emergent behaviors that are not easily predictable from individual agent capabilities alone.

\subsubsection{Predictive Capabilities}


The predictive capabilities enabled by world models represent a quantum leap beyond traditional approaches that focus primarily on reactive responses to current observations \cite{taniguchi2023world,batty2005modelling}. \textit{Multi-step look-ahead prediction} allows agents to forecast environment states across  time horizons, enabling planning and strategic decision-making~\cite{hafner2020mastering}. This predictive capability is particularly valuable in multi-agent systems where the effects of actions may not be immediately apparent but may have significant consequences over longer time periods.

\textit{Uncertainty quantification} represents another critical capability of advanced world models, as these systems must operate under significant uncertainty about other agents' intentions and capabilities \cite{bohm2019uncertainty,luo2018algorithmic,lockwood2022review,bi2023stochastic}. The models learn to represent both epistemic uncertainty, which arises from incomplete knowledge about the system, and aleatoric uncertainty, which is inherent in the stochastic nature of multi-agent interactions. This uncertainty quantification enables agents to make robust decisions that account for the inherent unpredictability of multi-agent systems.

\textit{Counterfactual environment generation} enables agents to explore alternative scenarios and evaluate different possible outcomes without actually executing actions in the real environment \cite{chen2023adversarial,madaan2021generate,parvaneh2020counterfactual,sauer2021counterfactual,kenny2021generating,fu2020counterfactual}. This capability is particularly valuable for planning and strategy development, as agents can explore the consequences of different action sequences and coordination strategies. The ability to generate realistic counterfactual scenarios enables agents to learn from simulated experiences and develop robust strategies that work well across a range of possible futures.

\subsection{Action Policy via Generative Modeling}

\subsubsection{Generative Policy Architecture}


The reconceptualization of action policies as generative models represents a fundamental shift from traditional policy representations that map observations directly to actions. Instead of learning simple input-output mappings, generative policies learn to model the complex distribution of optimal actions conditioned on the full context of multi-agent interactions. This approach enables agents to generate sophisticated action sequences that account for temporal and spatial dependencies, strategic interactions, and coordination requirements.

\textit{Generative policy architecture }treats multi-agent coordination as a sequence generation problem, where agents generate action sequences that are temporally consistent and strategically coherent~\cite{chi2023diffusion}. This sequence-to-sequence approach enables agents to plan extended action sequences that account for the anticipated responses of other agents and the evolving dynamics of the environment. The generative nature of these policies allows them to capture multi-modal action distributions, which is particularly important in multi-agent systems where there may be multiple equally valid ways to achieve coordination.

\textit{Contextual action generation} represents a sophisticated capability where policies generate actions that are conditioned on rich multi-agent context, including predictions of other agents' behaviors, estimates of their intentions and capabilities, and models of the evolving environment state. This contextual conditioning enables agents to adapt their actions dynamically to the specific multi-agent situation they face, rather than relying on fixed behavioral patterns that may not be appropriate for all situations.

\textit{Hierarchical policy generation} enables agents to operate at multiple levels of abstraction, from high-level strategic decisions about coordination strategies to low-level tactical decisions about specific actions~\cite{wang2023distributed}. This hierarchical approach mirrors the way humans approach complex coordination problems, where high-level goals and strategies guide lower-level action selection. The hierarchical structure also enables more efficient learning and better generalization, as agents can learn general coordination principles at high levels while adapting specific behaviors at lower levels.

\subsubsection{Communication and Coordination}


Communication and coordination capabilities emerge naturally from the generative approach to policy modeling \cite{krupnik2020multi,song2018multi,zhan2018generative}. Generative communication involves automatically generating relevant messages for coordination based on predicted future needs and current context. Rather than relying on predefined communication protocols, agents learn to generate messages that convey the information most relevant for coordination in specific situations. This adaptive communication capability enables agents to share information efficiently and effectively, even in situations that were not anticipated during system design.

\textit{Protocol-free coordination} represents another significant advantage of the generative approach, as agents can learn implicit coordination patterns without requiring predefined coordination mechanisms or communication protocols. The generative models learn to recognize and generate coordination patterns that emerge naturally from the structure of the multi-agent task and the capabilities of the agents involved. This flexibility is particularly valuable in dynamic environments where predefined protocols may become inadequate or inappropriate as conditions change.

\textit{Adaptive strategy generation} enables agents to create coordination strategies dynamically based on current multi-agent context and predicted future scenarios. Rather than relying on a fixed repertoire of coordination strategies, agents can generate new strategies that are tailored to specific situations. This capability is particularly important in complex, dynamic environments where the optimal coordination strategy may depend on subtle aspects of the current situation that cannot be captured by simple rules or predefined strategies.

{\em Related work on multi-agent RL  with communication.} Information sharing has shown to be crucial for   multi-agent coordination and performance improvement \cite{zhang2021multi,foerster2016learning,sukhbaatar2016learning}. In general, the related works aim to address questions ``what information to share'' and ``whom to communicate with''. Recent works \cite{jiang2018learning,foerster2016learning} adopt an end-to-end message-generation network to generate messages by encoding the past and current observation information. CommNet \cite{sukhbaatar2016learning} aggregates all the agents’ hidden states as the global message. Under partial observable environments, the aforementioned works do not capture future information, which can be essential for planning tasks. In this regard, MACI \cite{pretorius2020learning} allows agents to share the their imagined trajectories to other agents through world model rollout. Furthermore, \cite{kim2020communication} compresses the imagined trajectory into intention message to share with others. Notably, these works often require the information to be shared among all agents in the network. To reduce the communication burden, \cite{jiang2018learning} and \cite{liu2020multi} use the attention unit to select a group of collaborator to communicate while learning (or planning) directly in the (potentially high-dimensional) space. \cite{egorov2022scalable} considers the notion of ``locality'' where the agent receives history information from its neighbors in the environment.

\subsection{Integrated Prediction and Planning}

\subsubsection{Multi-Agent Predictive Planning}

The integration of world models and generative policies creates a comprehensive framework for multi-agent decision-making that combines sophisticated environmental modeling with strategic action generation. This integrated approach enables agents to engage in multi-agent predictive planning, where they use their learned models of environment dynamics and other agents' behaviors to plan action sequences that achieve their objectives while accounting for the complex interactions inherent in multi-agent systems.

Multi-agent predictive planning \cite{toumieh2022decentralized,rhinehart2019precog,negenborn2009multi,sun2022multi,gmytrasiewicz2005framework,luo2023jfp,zhao2019multi} involves \textit{distributed planning} where each agent maintains consistent generative world models while optimizing for local objectives that contribute to overall system performance. This distributed approach enables scalability while maintaining the benefits of sophisticated modeling and prediction. Each agent uses its world model to predict the consequences of different action sequences, both for itself and for other agents, enabling strategic decision-making that accounts for the full complexity of multi-agent interactions.

\textit{Hierarchical planning }\cite{dempster1981analytical,hafner2022deep,corkill1979hierarchical,chen2024simple,holler2020hddl,mohr2018ml,ozdamar1999hierarchical,li2022hierarchical,pateria2021hierarchical,naveed2021trajectory,wen2020efficiency} represents a sophisticated approach where agents plan at multiple levels of abstraction, from high-level coordination strategies to detailed action sequences. This hierarchical structure enables agents to maintain consistency between long-term strategic goals and short-term tactical decisions. High-level planning focuses on coordination strategies and role allocation, while low-level planning focuses on specific action sequences that implement these high-level strategies.

\subsubsection{Proactive Decision-Making Architecture}


The proactive decision-making architecture represents the culmination of the generative approach, enabling agents to anticipate and prepare for future multi-agent scenarios rather than simply reacting to current observations. Anticipatory systems \cite{muccini2019machine,wang2018proactive,moreno2016efficient} predict and prepare for future multi-agent scenarios through sophisticated modeling of other agents' behaviors, environment evolution, and emergent collective behaviors. Other-agent modeling involves generating detailed predictions of teammates' and opponents' behaviors based on learned generative models that capture their capabilities, preferences, and likely strategies.

\textit{Environment evolution modeling }\cite{niv2002evolution,padakandla2021survey,co2021evolving,moriarty1999evolutionary,bai2023evolutionary,ha2018recurrent} enables agents to anticipate how multi-agent interactions will shape the environment over time, accounting for both direct effects of agent actions and indirect consequences that may emerge from complex interaction patterns. This predictive capability enables agents to make decisions that account for long-term environmental changes rather than focusing only on immediate effects.

\textit{Emergent behavior prediction} \cite{martinez2017emergent,ndousse2021emergent,grupen2022concept,decker2016creatures,wu2017emergent,sharma2020emergent} represents perhaps the most sophisticated capability of the integrated framework, as agents learn to forecast collective system behaviors that emerge from individual agent interactions. This capability enables agents to anticipate and influence system-level outcomes rather than focusing only on individual performance. The ability to predict emergent behaviors is particularly valuable in large-scale multi-agent systems where collective behaviors may be qualitatively different from individual agent behaviors.

\textit{Strategic reasoning capabilities} \cite{zhang2024llm,duan2024gtbench,guo2025deepseek} enable decision-making that considers long-term multi-agent game dynamics and strategic interactions rather than myopic optimization. Agents learn to think strategically about their interactions with other agents, considering how their actions will influence other agents' future decisions and how these interactions will shape the long-term evolution of the system.

\textit{Adaptive coordination }\cite{li2014adaptive,entin1999adaptive,zhang2014leader,wan2010adaptive,crawford1995adaptive} represents the dynamic adjustment of coordination strategies based on predictions and changing circumstances. Anticipatory communication involves generating and sharing relevant information before it is explicitly requested, based on predictions of what information will be needed for effective coordination. Role adaptation enables agents to proactively adjust their roles and responsibilities based on predicted future needs and changing circumstances. Conflict prevention involves identifying and resolving potential conflicts before they manifest, based on predictions of how different action sequences might lead to conflicts or coordination failures.




\section{Research Directions and Implementation Pathways}

The realization of GenAI-based multi-agent reinforcement learning requires addressing several critical technical challenges while simultaneously exploring promising implementation pathways that can demonstrate the practical value of this approach. The research directions we outline represent both immediate opportunities for advancing the field and longer-term challenges that will require sustained effort from the research community.

\subsection{Technical Challenges and Discussions}


\textit{Scalability} represents one of the most fundamental challenges in developing practical GenAI-based multi-agent systems \cite{wang2023overview,bandi2023power,vadisetty2024generative}. The computational requirements for training and running generative models can be substantial, and these requirements typically scale unfavorably with the number of agents and the complexity of the environment. Developing efficient generative models for large-scale multi-agent systems requires advances in model architecture design, training algorithms, and computational optimization. Recent advances in transformer architectures and diffusion models~\cite{ren2024diffusion} provide promising foundations, but significant effort is needed to adapt these techniques for multi-agent scenarios where the models must capture complex interaction patterns between potentially hundreds or thousands of agents.

The scalability challenge extends beyond simple computational complexity to include fundamental questions about how to structure generative models to capture multi-agent interactions effectively. Traditional approaches that attempt to model all agents simultaneously quickly become intractable \cite{zhang2021multi,shoham2008multiagent,busoniu2008comprehensive}, while approaches that model agents independently fail to capture critical interaction patterns. New architectural innovations are needed that can capture multi-agent interactions efficiently while scaling to large numbers of agents.

\textit{Training stability} represents another critical challenge, particularly in multi-agent environments where multiple generative models are simultaneously adapting and evolving. The non-stationary nature of multi-agent environments creates fundamental challenges for training generative models, as the data distribution is constantly changing as agents adapt their behaviors. Ensuring stable learning requires developing new training algorithms that can handle the complex feedback loops and interdependencies that characterize multi-agent learning.

The training stability problem is compounded by the fact that generative models typically require large amounts of data and extensive training to achieve good performance. In multi-agent environments, generating sufficient training data can be challenging, particularly for rare but important scenarios such as coordination failures or conflict situations. New approaches are needed that can enable stable and efficient training of generative models in non-stationary multi-agent environments.

\textit{Evaluation metrics} represent a significant challenge in assessing the performance of GenAI-based multi-agent systems. Traditional metrics that focus on individual agent performance are inadequate for evaluating systems where the primary value comes from coordination and collective intelligence. Developing comprehensive benchmarks and evaluation metrics for generative multi-agent capabilities requires new approaches that can capture both individual agent performance and emergent collective behaviors \cite{brockman2016openai}.

The evaluation challenge is particularly complex because the value of generative capabilities may not be immediately apparent in simple metrics. For example, the ability to predict other agents' behaviors may not directly improve immediate task performance but may enable better long-term coordination and strategic planning. New evaluation frameworks are needed that can assess these sophisticated capabilities and their contribution to overall system performance.

\subsection{Applications and Use Cases}


The practical applications of GenAI-based multi-agent reinforcement learning span numerous domains where intelligent coordination and collaboration are critical for success \cite{feuerriegel2024generative}. These applications provide both motivation for developing the technology and testbeds for evaluating its effectiveness in real-world scenarios \cite{ramdurai2023impact,fui2023generative,law2024application,sengar2024generative,gozalo2023survey,weisz2024design}. In the following, we provide a few outstanding example applciations. 

\textit{Autonomous vehicle} coordination represents one of the most compelling and immediate applications of this technology \cite{gao2024cardreamer,xu2023generative}. Intelligent  transportation systems require sophisticated coordination between  vehicles to achieve safe and efficient traffic flow. Traditional approaches to autonomous vehicle coordination rely primarily on reactive behaviors and simple communication protocols that are inadequate for the complex coordination required in dense traffic scenarios. Proactive traffic management enabled by generative models can anticipate traffic patterns, predict potential conflicts, and coordinate vehicle behaviors to optimize both safety and efficiency.

The generative approach enables autonomous vehicles to predict the intentions and likely behaviors of other vehicles, pedestrians, and other traffic participants \cite{hu2023gaia,zheng2024genad}. This predictive capability allows vehicles to plan their actions proactively rather than simply reacting to immediate observations. For example, a vehicle can anticipate that another vehicle is likely to change lanes based on its current trajectory and positioning, enabling the first vehicle to adjust its behavior preemptively to facilitate safe and efficient coordination.

Collision avoidance represents a critical safety application where the predictive capabilities of generative models can provide significant advantages over traditional reactive approaches. By predicting potential conflict situations before they develop, generative-RL agents can take preventive actions that avoid dangerous situations entirely rather than relying on last-minute emergency responses \cite{marathe2023wedge}. This proactive approach to safety can significantly reduce accident rates and improve overall transportation system performance.

\textit{Swarm robotics} \cite{schranz2020swarm,navarro2013introduction,dias2021swarm} provides another compelling application domain where coordinated exploration and task execution require sophisticated coordination between large numbers of agents. Traditional approaches to swarm coordination rely on simple behavioral rules that can produce effective coordination in some scenarios but often fail in complex or dynamic environments. Generative models enable more sophisticated coordination strategies that can adapt to changing conditions and handle complex task requirements \cite{mandlekar2021matters}.

In swarm robotics applications, individual robots can use generative models to predict the behaviors of other robots and coordinate their actions accordingly \cite{khaldi2015overview,csahin2004swarm,tan2013research,dorigo2021swarm}. This capability enables more efficient exploration strategies where robots can avoid redundant coverage while ensuring comprehensive exploration of the environment. The predictive capabilities also enable more effective task allocation and load balancing, as robots can anticipate resource requirements and coordinate their efforts to achieve optimal overall performance \cite{chung2018survey}.

\textit{Multi-agent games} provide a controlled environment for developing and testing sophisticated strategic reasoning capabilities. Competitive and cooperative game scenarios require agents to model opponents' strategies, predict their likely actions, and develop counter-strategies that account for the dynamic nature of strategic interactions. Generative models enable agents to explore different strategic scenarios and develop robust strategies that perform well against a variety of opponents \cite{weisz2023toward,ferrara2024genai}.

The game domain also provides valuable insights into the strategic reasoning capabilities that are needed for real-world applications \cite{silver2016mastering,silver2018general,mnih2013playing}. The ability to model opponents' strategies and predict their likely responses is critical not only for game playing but also for any application where agents must interact with other intelligent agents that may have conflicting or partially aligned objectives.

\subsection{Future Opportunities}


The long-term potential of GenAI-based multi-agent reinforcement learning extends far beyond current applications to encompass transformative possibilities that could fundamentally change how we approach distributed intelligence and human-AI collaboration.

\textit{Human-agent collaboration} represents one of the most exciting and challenging opportunities for future development \cite{fan2008influence,cila2022designing,bradshaw2017human,fan2005extending,}. As AI systems become more capable and ubiquitous, the ability to collaborate effectively with human users becomes increasingly important. Generative models for predicting and adapting to human behavior can enable AI agents to work more effectively with human partners by anticipating human needs, adapting to human preferences, and providing appropriate assistance without being intrusive or disruptive.

The challenge of human-agent collaboration is complicated by the fact that humans have complex, often unpredictable behaviors and preferences that can be difficult to model accurately \cite{ramchurn2016human,kolling2015human,mukherjee2022survey,xing2021toward}. Generative models offer promising approaches for learning from human behavior data to predict likely human actions and preferences in different situations. These models can enable AI agents to adapt their behavior to work effectively with specific human partners while maintaining the flexibility to work with different humans with different preferences and capabilities.

\textit{Emergent communication} represents another fascinating opportunity where agents can automatically develop communication protocols and languages that are optimized for specific coordination tasks \cite{lazaridou2020emergent,gupta2020networked,eccles2019biases}. Rather than relying on predefined communication protocols, agents can learn to develop their own communication systems that are tailored to their specific coordination requirements. This capability could enable more efficient and effective communication in scenarios where traditional communication protocols are inadequate or inappropriate. The development of emergent communication systems requires sophisticated generative models that can learn to associate communication signals with coordination outcomes and develop communication strategies that maximize coordination effectiveness \cite{lee2017emergent,ren2010distributed}. This research direction connects to fundamental questions in linguistics and cognitive science about how communication systems develop and evolve.

\textit{Meta-learning} capabilities represent another ambitious long-term opportunity, enabling agents to rapidly adapt to new multi-agent environments and tasks through few-shot learning approaches \cite{ankile2024imitation}. Rather than requiring extensive training for each new task or environment, meta-learning enabled agents could quickly adapt their behaviors based on limited experience with new situations. This capability would dramatically expand the practical applicability of multi-agent systems by reducing the time and data requirements for deploying agents in new domains. The meta-learning challenge requires developing generative models that can capture general principles of multi-agent coordination and adaptation that transfer across different domains and tasks. This requires understanding the fundamental patterns and structures that underlie effective multi-agent coordination and developing models that can recognize and exploit these patterns in new situations.

\section{Conclusion and Call to Action}

The paradigm shift from reactive to proactive multi-agent intelligence represents a promising approach to distributed AI systems that has the potential to transform how we design, deploy, and interact with intelligent agents in complex environments. By reconceptualizing agents as generative models capable of synthesizing behaviors, predicting interactions, and making anticipatory decisions, we can address long-standing challenges in multi-agent learning while opening new possibilities for collective intelligence that transcends the capabilities of individual agents. In general, GenAI-based approaches to MARL offer great promise with tremendous opportunities and yet many open problems, and this article touches only the tip of iceberg.

\subsection{Transformative Potential}


The transformative potential of this approach extends across multiple dimensions of distributed AI systems, from fundamental theoretical advances to practical applications that can address critical societal challenges. The implications for distributed AI systems include enabling more sophisticated coordination and collaboration mechanisms that can handle the complexity and uncertainty inherent in real-world multi-agent environments. Traditional approaches that rely on simple coordination mechanisms and reactive behaviors are fundamentally limited in their ability to handle complex, dynamic environments where optimal coordination strategies may depend on subtle patterns of interaction and prediction.

The advancement toward collective intelligence represents a qualitative leap beyond systems that simply coordinate individual agents toward truly intelligent collective behaviors that emerge from sophisticated interaction patterns. This collective intelligence can enable multi-agent systems to solve problems that are beyond the capabilities of individual agents, even very capable ones. The emergent properties of well-coordinated multi-agent systems can produce solutions and behaviors that are not easily predictable from individual agent capabilities alone.

Real-world applications across autonomous systems, robotics, smart cities, and other domains requiring intelligent coordination stand to benefit significantly from these advances. The ability to deploy sophisticated multi-agent systems that can adapt to changing conditions, coordinate effectively with minimal communication overhead, and handle complex, dynamic environments opens possibilities for addressing some of the most challenging problems facing society today.

\subsection{Research Priorities}


The realization of this vision requires focused effort on several key technical challenges that represent both immediate opportunities and longer-term research priorities. Developing rigorous theoretical frameworks for understanding generative multi-agent learning and its convergence properties is essential for establishing the scientific foundation for this field. Current theoretical understanding of multi-agent learning is largely based on traditional approaches that may not apply to generative models, and new theoretical frameworks are needed to understand the behavior and properties of generative multi-agent systems.

Creating efficient generative model architectures that can handle large-scale multi-agent systems represents a critical engineering challenge that requires innovations in both model design and computational optimization. The computational requirements for training and running generative models can be substantial, and new approaches are needed to make these systems practical for large-scale deployment.

Establishing comprehensive benchmarks and metrics for assessing generative multi-agent capabilities is essential for enabling systematic progress in the field \cite{2022marllib}. Current evaluation approaches are inadequate for assessing the sophisticated capabilities enabled by generative models, and new evaluation frameworks are needed that can capture both individual agent performance and emergent collective behaviors.

Ensuring that generative agents operate safely and reliably in critical applications represents perhaps the most important long-term challenge for the field. As these systems become more capable and are deployed in safety-critical applications, it becomes essential to understand their failure modes and develop approaches for ensuring reliable operation even in unexpected or adversarial conditions.

\subsection{Roadmap to Implementation}


The path toward realizing GenAI-based multi-agent intelligence requires a coordinated research and development effort that builds systematically from foundational advances toward practical applications. \textbf{The short-term milestones} for the next one to two years should focus on developing foundational world model architectures specifically designed for multi-agent systems. This includes adapting recent advances in generative modeling, such as diffusion models and transformer architectures, to capture the complex interaction patterns that characterize multi-agent environments. These foundational models must be able to learn compact representations of multi-agent dynamics while maintaining the expressiveness needed to capture subtle coordination patterns and strategic interactions.

The implementation of generative communication protocols represents another critical short-term milestone that can demonstrate the practical value of the generative approach. These protocols should enable agents to automatically generate and interpret messages that convey complex coordination information efficiently. Unlike traditional communication protocols that rely on predefined message formats, generative communication protocols can adapt dynamically to the specific coordination requirements of different situations.

Demonstration of proactive decision-making capabilities in controlled environments provides essential validation of the core concepts underlying the generative approach. These demonstrations should clearly show the advantages of proactive planning and prediction over traditional reactive approaches in multi-agent coordination tasks. The controlled environments should be sophisticated enough to capture the essential challenges of multi-agent coordination while being simple enough to enable clear evaluation and comparison of different approaches.

\textbf{Medium-term goals} spanning the next three to five years should focus on scaling these foundational capabilities toward real-world deployment and integration with human users. Deployment in real-world applications such as autonomous vehicle coordination represents a critical test of the practical viability of the approach. These deployments must demonstrate that generative multi-agent systems can operate safely and effectively in complex, dynamic environments with real-world constraints and uncertainties.

The integration of human-agent collaboration capabilities represents another important medium-term goal that can significantly expand the practical applicability of multi-agent systems. This includes developing generative models that can predict and adapt to human behavior patterns, enabling AI agents to work effectively with human partners in collaborative tasks. The challenge of human-agent collaboration requires understanding not only human behavioral patterns but also human preferences, expectations, and communication styles.

Development of meta-learning frameworks for rapid adaptation represents a sophisticated capability that can enable multi-agent systems to quickly adapt to new environments and tasks. This capability is particularly important for practical deployment, as it reduces the time and data requirements for adapting multi-agent systems to new applications. Meta-learning frameworks must capture general principles of multi-agent coordination that can transfer across different domains while maintaining the flexibility to adapt to domain-specific requirements \cite{kim24openvla,liu2024rdt,li2024cogact}.

\textbf{The long-term vision} extending beyond five years encompasses large-scale deployment of generative multi-agent systems across multiple application domains, the emergence of truly collective intelligence in distributed systems, and integration across various sectors of society and the economy. Large-scale deployment requires not only technical advances but also the development of appropriate regulatory frameworks, safety standards, and ethical guidelines for deploying sophisticated AI systems in critical applications.

The emergence of truly collective intelligence represents a qualitative transformation where multi-agent systems exhibit intelligent behaviors that are genuinely collective rather than simply the sum of individual agent capabilities. This collective intelligence could enable solutions to complex problems that require coordination and collaboration across multiple scales and domains, from local coordination between small groups of agents to global coordination between large-scale distributed systems.

Integration across multiple application domains requires developing generative multi-agent frameworks that are general enough to apply across different domains while being flexible enough to adapt to domain-specific requirements. This integration could enable synergies between different applications and create opportunities for transferring insights and capabilities across domains.

The success of this vision requires collaborative effort across the research community, bringing together expertise in generative AI, reinforcement learning, multi-agent systems, and application domains. The interdisciplinary nature of this challenge means that progress will require not only technical advances but also new forms of collaboration that can bridge different research communities and application domains. By working together toward these ambitious goals, the research community can realize the transformative potential of generative multi-agent intelligence and create the foundation for next-generation distributed AI systems that can address some of the most challenging problems facing society.

The implementation pathway also requires careful attention to ethical considerations and societal impacts as these powerful technologies are developed and deployed. As generative multi-agent systems become more capable and ubiquitous, it becomes increasingly important to ensure that they are developed and deployed in ways that benefit society while minimizing potential risks and negative consequences. This requires ongoing dialogue between researchers, policymakers, and society to ensure that the development of these technologies aligns with human values and societal needs.

The roadmap we have outlined represents an ambitious but achievable path toward realizing the transformative potential of GenAI-based multi-agent reinforcement learning. Success will require sustained effort, significant resources, and collaboration across multiple research communities, but the potential benefits for advancing distributed intelligence and addressing critical societal challenges make this effort both worthwhile and necessary. The time is right for this paradigm shift, as recent advances in generative AI provide the foundational technologies needed to make this vision a reality, while growing societal needs for intelligent coordination and collaboration create strong motivation for developing these capabilities.

\newpage 

 \printbibliography

\end{document}